\documentclass[runningheads]{llncs}
\usepackage[breaklinks=true,colorlinks,bookmarks=false]{hyperref}
\usepackage{amssymb}
\usepackage{amsmath}
\usepackage{amsmath,amssymb}
\usepackage{psfrag}
\usepackage{epsfig}
\usepackage{cite}
\usepackage{color}
\usepackage{subfigure}
\usepackage{multirow}
\usepackage{threeparttable}
\usepackage{booktabs}
\usepackage{mmstyles}
\usepackage{marvosym}
\usepackage{bbm}
\usepackage[T1]{fontenc}

\begin{document}
%

\title{TCEIP: Text Condition Embedded Regression Network for Dental Implant Position Prediction}
\titlerunning{TCEIP}
%
%


\author{
Xinquan Yang\inst{1,2,3},
Jinheng Xie\inst{1,2,3,5},
Xuguang Li\inst{4} ,
Xuechen Li\inst{1,2,3},
Xin Li\inst{4},
Linlin Shen\inst{1,2,3}\Letter,
Yongqiang Deng\inst{4}
}

\institute{College of Computer Science and Software Engineering, Shenzhen University, Shenzhen, China \\ \and AI Research Center for Medical Image Analysis and Diagnosis, Shenzhen University, Shenzhen, China \and National Engineering Laboratory for Big Data System Computing Technology, Shenzhen University, China \and Department of Stomatology, Shenzhen University General Hospital, Shenzhen, China \and National University of Singapore, Singapore\\
\email{yangxinquan2021@email.szu.edu.cn, llshen@szu.edu.cn}
}


%
\maketitle              
\begin{abstract}
When deep neural network has been proposed to assist the dentist in designing the location of dental implant, most of them are targeting simple cases where only one missing tooth is available. As a result, literature works do not work well when there are multiple missing teeth and easily generate false predictions when the teeth are sparsely distributed. In this paper, we are trying to integrate a weak supervision text, the target region, to the implant position regression network, to address above issues. We propose a text condition embedded implant position regression network (TCEIP), to embed the text condition into the encoder-decoder framework for improvement of the regression performance. A cross-modal interaction that consists of cross-modal attention (CMA) and knowledge alignment module (KAM) is proposed to facilitate the interaction between features of images and texts. The CMA module performs a cross-attention between the image feature and the text condition, and the KAM mitigates the knowledge gap between the image feature and the image encoder of the CLIP. Extensive experiments on a dental implant dataset through five-fold cross-validation demonstrated that the proposed TCEIP achieves superior performance than existing methods.

\keywords{Dental Implant \and Deep Learning \and Text Guided Detection \and Cross-Modal Interaction}
\end{abstract}

\section{Introduction}
According to a systematic research study~\cite{elani2018trends}, periodontal disease is the world's 11th most prevalent oral condition, which potentially causes tooth loss in adults, especially the aged~\cite{nazir2020global}. One of the most appropriate treatments for such a defect/dentition loss is prosthesis implanting, in which the surgical guide is usually used. However, dentists must load the Cone-beam computed tomography (CBCT) data into the surgical guide design software to estimate the implant position, which is tedious and inefficient. In contrast, deep learning-based methods show great potential to efficiently assist the dentist in locating the implant position~\cite{liu2021transfer}.

\begin{figure}
\centering
\includegraphics[width=1.0\linewidth]{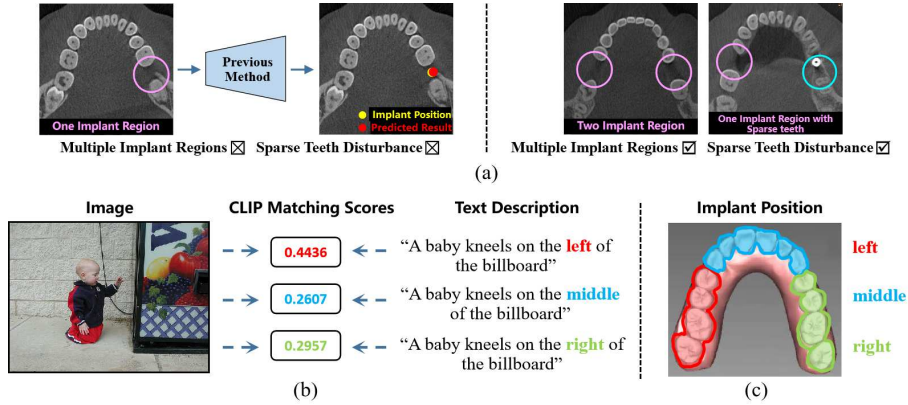}
\caption{(a) The 2D axial view of tooth crown images captured from different patients, where the pink and blue circles denote the implant and sparse teeth regions, respectively. (b) The matching score of the CLIP for a pair of image and text. (c) The division teeth region.} \label{fig_describe}
\end{figure}

Recently, deep learning-based methods have achieved great success in the task of implant position estimation. Kurt et al.~\cite{kurt2021deep} and Widiasri et al.~\cite{widiasri2022dental} utilized the convolutional neural network (CNN) to locate the oral bone, e.g., the alveolar bone, maxillary sinus and jaw bone, which determines the implant position indirectly. Different from these implant depth measuring methods, Yang et al.~\cite{yang2022implantformer} developed a transformer-based implant position regression network (ImplantFormer), which directly predicts the implant position on the 2D axial view of tooth crown images and projects the prediction results back to the tooth root by the space transform algorithm. However, these methods generally consider simple situations, in which only one missing tooth is available. When confronting some special cases, such as multiple missing teeth and sparse teeth disturbance in Fig.~\ref{fig_describe}(a), the above methods may fail to determine the correct implant position. In contrast, clinically, dentists have a subjective expertise about where the implant should be planted, which motivates us that, additional indications or conditions from dentists may help predict an accurate implant position.


In recent years, great success has been witnessed in Vision-Language Pre-training (VLP). For example, Radford~\cite{radford2021learning} proposed Contrastive Language-Image Pretraining (CLIP) to learn diverse visual concepts from 400 million image-text pairs automatically, which can be used for vision tasks like object detection~\cite{zhou2022detecting} and segmentation~\cite{xie2022clims}. In this paper, we found that CLIP has the ability to learn the position relationship among instances. We showcase examples in Fig.~\ref{fig_describe}(b) that the image-text pair with the word 'left' get a higher matching score than others, as the position of baby is on the left of the billboard.

Motivated by the above observation in dental implant and the property of CLIP, in this paper, we integrate a text condition from the CLIP to assist the implant position regression. According to the natural distribution, we divide teeth regions into three categories in Fig.~\ref{fig_describe}(c), i.e., left, middle, and right. Specifically, during training, one of the text prompts, i.e., 'right', 'middle', and 'left' is paired with the crown image as input, in which the text prompt works as a guidance or condition. The crown image is processed by an encoder-decoder network for final location regression. In addition, to facilitate the interaction between features in two modalities, a cross-modal interaction that consists of cross-modal attention (CMA) and knowledge alignment module (KAM), is devised. The CMA module fuses conditional information, i.e., text prompt, to the encoder-decoder. This brings additional indications or conditions from the dentist to help the implant position regression. However, a knowledge gap may exist between our encoder-decoder and CLIP. To mitigate the problem, the KAM is proposed to distill the encoded-decoded features of crown images to the space of CLIP, which brings significant localization improvements. In inference, given an image, the dentist just simply gives a conditioning text like "let's implant a prosthesis on the left", the network will preferentially seek a suitable location on the left for implant prosthesis.

Main contributions of this paper can be summarized as follows: 1) To the best of our knowledge, the proposed TCEIP is the first text condition embedded implant position regression network that integrates a text embedding of CLIP to guide the prediction of implant position. (2) A cross-modal interaction that consists of a cross-modal attention (CMA) and knowledge alignment module (KAM) is devised to facilitate the interaction between features that representing image and text. (3) Extensive experiments on a dental implant dataset demonstrated the proposed TCEIP achieves superior performance than the existing methods, especially for patients with multiple missing teeth or sparse teeth.

\section{Method}
\begin{figure}[!ht]
\centering
\includegraphics[width=1.0\linewidth]{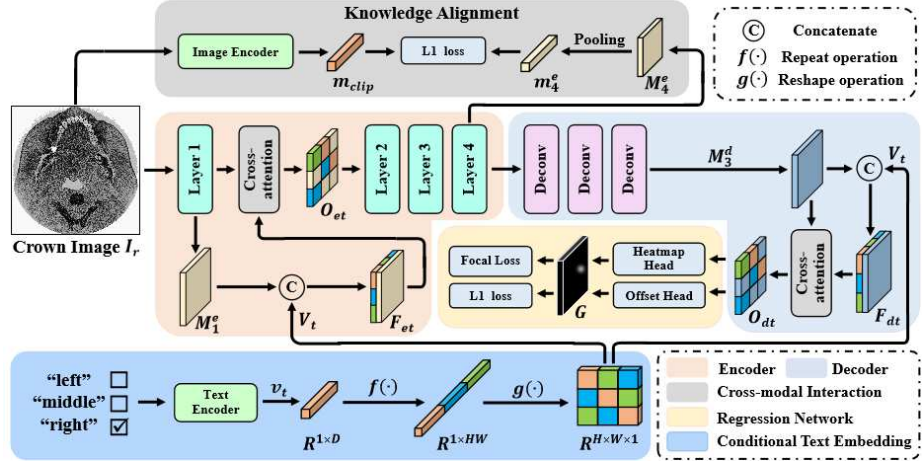}
\caption{The network architecture of the proposed prediction network.} \label{fig2}
\end{figure}

Given a tooth crown image with single or multiple implant regions, the proposed TCEIP aims to give a precise implant location conditioned by text indications from the dentist, i.e., a description of position like 'left', 'right', or 'middle'. An overview of TCEIP is presented in Fig.~\ref{fig2}. It mainly consists of four parts: i) Encoder and Decoder, ii) Conditional Text Embedding, iii) Cross-Modal Interaction Module, and iv) Heatmap Regression Network. After obtaining the predicted coordinates of the implant at the tooth crown, we adopt the space transformation algorithm~\cite{yang2022implantformer} to fit a centerline of implant to project the coordinates to the tooth root, where the real implant location can be acquired. Next, we will introduce these modules in detail.

\subsection{Encoder and Decoder}
We employ the widely used ResNet~\cite{he2016deep} as the encoder of TCEIP. It mainly consists of four layers and each layer contains multiple residual blocks. Given a tooth crown image $\mI_r$, a set of feature maps, i.e., $\{\mM^e_1, \mM^e_2, \mM^e_3, \mM^e_4\}$, can be accordingly extracted by the ResNet layers. Each feature map has a spatial and channel dimension. To ensure fine-grained heatmap regression, three deconvolution layers are adopted as the Decoder to recover high-resolution features. It consecutively upsamples feature map $\mM^e_4$ as high-resolution feature representations, in which a set of recovered features $\{\mM^d_1, \mM^d_2, \mM^d_3\}$ can be extracted. Feature maps $\mM_1^e$,  $\mM_4^e$ and $\mM_3^d$ will be further employed in the proposed modules, where $\mM_1^e$ and $\mM_3^d$ have the same spatial dimension $\mathbb{R}^{128\times 128\times C}$ and $\mM_4^e\in \mathbb{R}^{16\times 16\times \hat C}$.

\subsection{Conditional Text Embedding}
To integrate the text condition provided by a dentist, we utilize the CLIP to extract the text embedding. Specifically, additional input of text, e.g., 'left', 'middle', or 'right', is processed by the CLIP Text Encoder to obtain a conditional text embedding $\vv_t \in \mathbb{R}^{1\times D}$. As shown in Fig.~\ref{fig2}, to interact with the image features from ResNet layers, a series of transformation $f(\cdot)$ and $g(\cdot)$ over $\vv_t$ are performed as follow:
\begin{equation}
\mV_t=g(f(\vv_t)) \in \mathbb{R}^{H\times W\times 1},
\end{equation}
where $f(\cdot)$ repeats text embedding $\vv_t$ from $\mathbb{R}^{1\times D}$ to $\mathbb{R}^{1\times HW}$ and $g(\cdot)$ then reshapes it to $\mathbb{R}^{H\times W\times 1}$. This operation ensures better interaction between image and text in the same feature space.

\begin{figure}[h]
\centering
\includegraphics[width=0.7\linewidth]{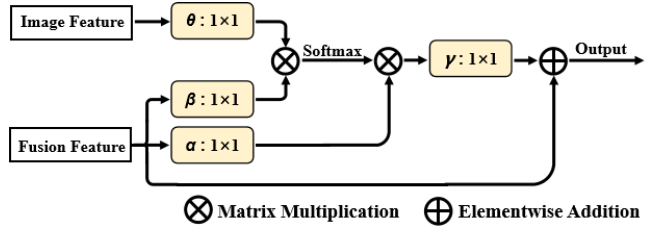}
\caption{The architecture of the proposed cross-modal attention module.} \label{fig3}
\end{figure}

\subsection{Cross-Modal Interaction}
High-resolution features from the aforementioned decoder can be directly used to regress the implant position. However, it cannot work well in situations of multiple teeth loss or sparse teeth disturbance. In addition, although we have extracted the conditional text embedding from the CLIP to assist the network regression, there exists a big difference with the feature of encoder-decoder in the feature space. To tackle these issues, we propose cross-modal interaction, including i) Cross-Modal Attention and ii) Knowledge Alignment module, to integrate the text condition provided by the dentist.


\subsubsection{Cross-Modal Attention Module} To enable the transformed text embedding $\mV_t$ better interact with intermediate features of the encoder and decoder, we design and plug a cross-modal attention (CMA) module into the shallow layers of the encoder and the final deconvolution layer. The architecture of CMA is illustrated in Fig.~\ref{fig3}. Specifically, the CMA module creates cross-attention between image features $\mM_1^e$ and fusion feature $\mF_{et} = [\mM_1^e|\mV_t]$ in the encoder, and image features $\mM_3^d$ and fusion feature $\mF_{dt} = [\mM_3^d|\mV_t]$ in the decoder, where $\mF_{et}, \mF_{dt} \in \mathbb{R}^{H\times W\times (C+1)}$. The CMA module can be formulated as follows:
\begin{equation}
\mO=\gamma(\text{Softmax}(\theta(\mM) \beta(\mF))\alpha(\mF)) + \mF,
\end{equation}
where four independent 1$\times$1 convolutions $\alpha, \beta, \theta$, and $\gamma$ are used to map image and fusion features to the space for cross-modal attention. At first, $\mM$ and $\mF$ are passed into $\theta(\cdot)$, $\beta(\cdot)$ and $\alpha(\cdot)$ for channel transformation, respectively. Following the transformed feature, $\mM_\theta$ and $\mF_\beta$ perform multiplication via a Softmax activation function to take a cross-attention with $\mF_\alpha$. In the end, the output feature of the cross-attention via $\gamma(\cdot)$ for feature smoothing is added with $\mF$. Given the above operations, the cross-modal features $\mO_{et}$ and $\mO_{dt}$ are obtained and passed to the next layer.

\subsubsection{Knowledge Alignment Module}
The above operations only consider the interaction between features in two modalities. A problem is that text embeddings from pre-trained text encoder of CLIP are not well aligned with the image features initialized by ImageNet pre-training. This knowledge shift potentially weakens the proposed cross-modal interaction to assist the prediction of implant position. To mitigate this problem, we propose the knowledge alignment module (KAM) to gradually align image features to the feature space of pre-trained CLIP. Motivated by knowledge distillation~\cite{rasheed2022bridging}, we formulate the proposed knowledge alignment as follows:
\begin{equation}
    \mathcal{L}_{align} = |\vm^e_4 - \vm_{clip}|,
\end{equation}
where $\vm^e_4\in \mathbb{R}^{1\times D}$ is the transformed feature of $\mM^e_4$ after attention pooling operation~\cite{dosovitskiy2020image} and dimension reduction with convolution. $\vm_{clip}\in \mathbb{R}^{1\times D}$ is the image embedding extracted by the CLIP Image Encoder. Using this criteria, the encoder of TCEIP approximates the CLIP image encoder and consequently aligns the image features of the encoder with the CLIP text embeddings.

\subsection{Heatmap Regression Network} The heatmap regression network is used for locating the implant position, which consists of the heatmap and the offset head. The output of the heatmap head is the center localization of implant position, which is formed as a heatmap $\mG\in [0,1]^{H\times W}$. Following~\cite{law2018cornernet}, given coordinate of the ground truth implant location $(\tilde{t}_x,\tilde{t}_y)$, we apply a 2D Gaussian kernel to get the target heatmap:
\begin{equation}
\mG_{xy}=\exp(-\frac{(x-{\tilde{t}_x})^2+(y-{\tilde{t}_y})^2}{2\sigma^2})
\end{equation}
where $\sigma$ is an object size-adaptive standard deviation. The predicted heatmap is optimized by the focal loss~\cite{lin2017focal}:
\begin{equation}
\mathcal{L}_h=\frac{-1}{N}\sum_{xy}\left\{
\begin{array}{ccl}
(1-\hat \mG_{xy})^\lambda\log(\hat \mG_{xy}) & \text{if $\mG_{xy}=1$} \\
(1-\hat \mG_{xy})^\varphi\log(\hat \mG_{xy})^\lambda\log(1-\hat \mG_{xy}) & \text{otherwise}
\end{array}\right.
\end{equation}
where $\lambda$ and $\varphi$ are the hyper-parameters of the focal loss, $\hat{\mG}$ is the predicted heatmap and $N$ is the number of implant annotation in image. The offset head computes the discretization error caused by the downsampling operation, which is used to further refine the predicted location. The local offset loss $\mathcal{L}_o$ is optimized by the L1 loss. The overall training loss of network is:
\begin{equation}
\mathcal{L}=\mathcal{L}_h+\mathcal{L}_o+\mathcal{L}_{align}
\end{equation}

\subsection{Coordinate Projection} The output of TCEIP is the coordinate of implant at the tooth crown. To obtain the real implant location at the tooth root, we fit a centerline of implant using the predicted implant position of TCEIP and then extend the centerline to the root area, which is identical as~\cite{yang2022implantformer}. By this means, the intersections of implant centerline with 2D slices of root image, i.e. the implant position at the tooth root area, can be obtained.

\begin{figure}
\centering
\includegraphics[width=0.85\linewidth]{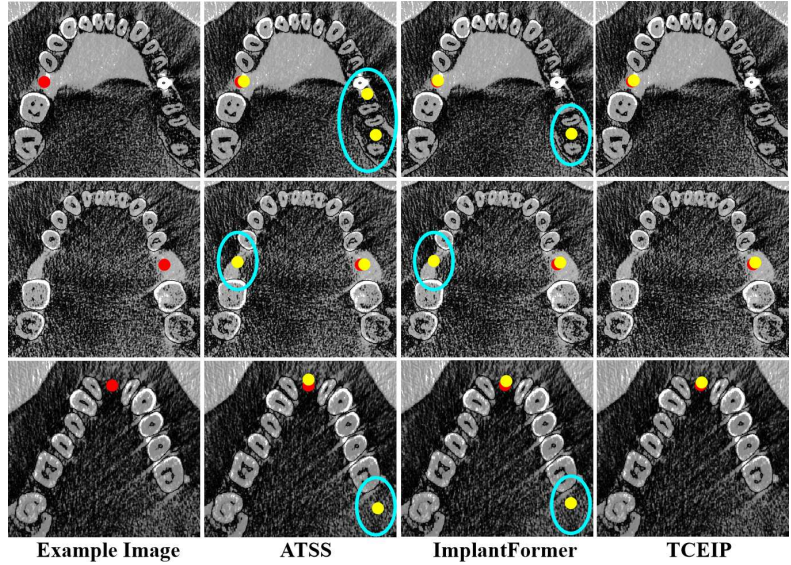}
\caption{Visual comparison of the predicted results with different detectors. The yellow and red circles represent the predicted implant position and ground-truth position, respectively. The blue ellipses denote false positive detections.} \label{fig4}
\end{figure}

\section{Experiments and Results}
\subsection{Dataset and Implementation Details}
The dental implant dataset was collected by~\cite{yang2022implantformer}, which contains 3045 2D slices of tooth crown images. The implant position annotations are annotated by three experienced dentists. The input image size of network is set as $512\times 512$. We use a batch size of 8, Adam optimizer and a learning rate of 0.001 for the network training. Total training epochs is 80 and the learning rate is divided by 10 when epoch $=\{40, 60\}$. The same data augmentation methods in~\cite{yang2022implantformer} was employed.

\begin{table}
\caption{The ablation experiments of each components in TCEIP.}\label{table1}
\centering
\scalebox{0.9}{
\begin{tabular}{c|c|c|c|c|c}
\toprule
Network                  & KAM          & Text Condition & Feature Fusion & CMA & $AP_{75}$\%     \\ \hline
\multirow{6}{*}{TCEIP} &              &             &              &              & 10.9$\pm$0.2457   \\
                         &              & \checkmark  &              &              & 14.6$\pm$0.4151 \\
                          & \checkmark  & \checkmark  &              &              & 15.7$\pm$0.3524 \\
                          & \checkmark  & \checkmark  &              & \checkmark & 16.5$\pm$0.3891 \\
                          & \checkmark  & \checkmark  & \checkmark &              & 17.1$\pm$0.2958 \\
                          & \checkmark  & \checkmark  & \checkmark & \checkmark & \textbf{17.8}$\pm$\textbf{0.3956} \\
\bottomrule
\end{tabular}
}
\end{table}

\begin{table}
\caption{Comparison of the proposed method with other mainstream detectors.}\label{table2}
\centering
\scalebox{0.9}{
\begin{tabular}{cccc}
\toprule
Methods     & Network         & Backbone                   & $AP_{75}\%$    \\ \hline
\multicolumn{1}{c}{\multirow{2}{*}{Transformer-based}} & ImplantFormer          & ViT-Base-ResNet-50                   & 13.7$\pm$0.2045 \\ \cline{3-3}
\multicolumn{1}{c}{}                                   & Deformable DETR~\cite{zhu2020deformable} &                            & 12.8$\pm$0.1417
\\ \cline{1-2}

\multicolumn{1}{c}{\multirow{5}{*}{CNN-based}}       & CenterNet~\cite{zhou2019objects}       &\multirow{5}{*}{ResNet-50}                         & 10.9$\pm$0.2457 \\
\multicolumn{1}{c}{}                                   & ATSS~\cite{zhang2020bridging}            &                            & 12.1$\pm$0.2694  \\
\multicolumn{1}{c}{}                                   & VFNet~\cite{zhang2021varifocalnet}           &                            & 11.8$\pm$0.8734 \\
\multicolumn{1}{c}{}                                   & RepPoints~\cite{yang2019reppoints}       &                            & 11.2$\pm$0.1858  \\ \cline{1-2}

\multicolumn{1}{c}{\multirow{1}{*}{-}}     & TCEIP   &    & \textbf{17.8}$\pm$\textbf{0.3956} \\
\bottomrule
\end{tabular}}
\end{table}

\subsection{Performance Analysis}
We use the same evaluation criteria in~\cite{yang2022implantformer}, i.e., average precision (AP) to evaluate the performance of our network. As high accurate position prediction is required in clinical practice, the IOU threshold is set as 0.75. Five-fold cross-validation was performed for all our experiments.

\subsubsection{Ablation Studies} To evaluate the effectiveness of the proposed network, we conduct ablation experiments to investigate the effect of each component in Table~\ref{table1}. We can observe from the second row of the table that the introduction of text condition improves the performance by 3.7\%, demonstrating the validity of using text condition to assist the implant position prediction. When combining both text condition and KAM, the improvement reaches 4.8\%. As shown in the table's last three rows, both feature fusion operation and CAM improve AP value by 1.4\% and 0.8\%, respectively. When combining all these components, the improvement reaches 6.9\%.

\subsubsection{Comparison to the Mainstream Detectors} To demonstrate the superior performance of the proposed TCEIP, we compare the AP value with the mainstream detectors in Table~\ref{table2}. Only the anchor-free detector is used for comparison, due to the reason that no useful texture is available around the center of the implant. As the teeth are missing, the anchor-based detectors can not regress the implant position successfully. From the table we can observe that, the transformer-based methods perform better than the CNN-based networks (e.g., ImplantFormer achieved 13.7\% AP, which is 1.6\% higher than the best-performed anchor-free network - ATSS). The proposed TCEIP achieves the best AP value - 17.8\%, among all benchmarks, which surpasses the ImplantFormer with a large gap. The experimental results proved the effectiveness of our method.

In Fig.~\ref{fig4}, we choose two best-performed detectors from the CNN-based (e.g., ATSS) and transformer-based (e.g., ImplantFormer) methods for visual comparison, to further demonstrate the superiority of TCEIP in the implant position prediction. The first row of the figure is a patient with sparse teeth, and the second and third rows are a patient with two missing teeth. We can observe from the figure that both the ATSS and ImplantFormer generate false positive detection, except for the TCEIP. Moreover, the implant position predicted by the TCEIP is more accurate. These visual results demonstrated the effectiveness of using text condition to assist the implant position prediction.

\section{Conclusions}
In this paper, we introduce TCEIP, a text condition embedded implant position regression network, which integrate additional condition from the CLIP to guide the prediction of implant position. A cross-modal attention (CMA) and knowledge alignment module (KAM) is devised to facilitate the interaction between features in two modalities. Extensive experiments on a dental implant dataset through five-fold cross-validation demonstrated that the proposed TCEIP achieves superior performance than the existing methods.

\bibliographystyle{splncs04}
\bibliography{ref}






\end{document}